\def\year{2022}\relax
\documentclass[letterpaper]{article} % DO NOT CHANGE THIS
\usepackage{aaai22}  % DO NOT CHANGE THIS
\usepackage{times}  % DO NOT CHANGE THIS
\usepackage{helvet}  % DO NOT CHANGE THIS
\usepackage{courier}  % DO NOT CHANGE THIS
\usepackage[hyphens]{url}  % DO NOT CHANGE THIS
\usepackage{graphicx} % DO NOT CHANGE THIS
\usepackage{natbib}  % DO NOT CHANGE THIS AND DO NOT ADD ANY OPTIONS TO IT
\usepackage{caption} % DO NOT CHANGE THIS AND DO NOT ADD ANY OPTIONS TO IT
\DeclareCaptionStyle{ruled}{labelfont=normalfont,labelsep=colon,strut=off} % DO NOT CHANGE THIS
\usepackage{algorithm}
\usepackage{algorithmic}

\usepackage{newfloat}
\usepackage{listings}
\floatstyle{ruled}
\newfloat{listing}{tb}{lst}{}
\floatname{listing}{Listing}
\title{Temporal Pyramid Transformer with Multimodal Interaction for\\ Video Question Answering}
\author{
    Min Peng,\textsuperscript{\rm 1,2}\thanks{Equal contribution. Under review.}
    Chongyang Wang,\textsuperscript{\rm 3}\footnotemark[1]
    Yuan Gao,\textsuperscript{\rm 4}
    Yu Shi,\textsuperscript{\rm 1}
    Xiang-Dong Zhou\textsuperscript{\rm 1}
    \\
}
\affiliations{
    \textsuperscript{\rm 1}Chongqing Institute of Green and Intelligent Technology, Chinese Academy of Sciences\\
    \textsuperscript{\rm 2}Chongqing School, University of Chinese Academy of Sciences\\
    \textsuperscript{\rm 3}University College London
    \textsuperscript{\rm 4}Shenzhen Institute of Artificial Intelligence and Robotics for Society\\
    pengmin@cigit.ac.cn, chongyang.wang.17@ucl.ac.uk, gaoyuankidult@gmail.com, \{shiyu;zhouxiangdong\}@cigit.ac.cn
}

\begin{document}

\maketitle

\begin{abstract}
Video question answering (VideoQA) is challenging given its multimodal combination of visual understanding and natural language understanding. While existing approaches seldom leverage the appearance-motion information in the video at multiple temporal scales, the interaction between the question and the visual information for textual semantics extraction is frequently ignored. Targeting these issues, this paper proposes a novel Temporal Pyramid Transformer (TPT) model with multimodal interaction for VideoQA. The TPT model comprises two modules, namely Question-specific Transformer (QT) and Visual Inference (VI). Given the temporal pyramid constructed from a video, QT builds the question semantics from the coarse-to-fine multimodal co-occurrence between each word and the visual content. Under the guidance of such question-specific semantics, VI infers the visual clues from the local-to-global multi-level interactions between the question and the video. Within each module, we introduce a multimodal attention mechanism to aid the extraction of question-video interactions, with residual connections adopted for the information passing across different levels. Through extensive experiments on three VideoQA datasets, we demonstrate better performances of the proposed method in comparison with the state-of-the-arts. \textit{Code available at https://github.com/Trunpm/TPT-for-VideoQA}
\end{abstract}

\section{Introduction}
With the advances of deep learning in computer vision and natural language processing \cite{REF1,REF2,REF3}, Video Question Answering (VideoQA) receives increasing attention recently, for its broad application in video retrieval, intelligent QA system and automatic drive. In comparison with image question answering \cite{REF4,REF5,REF6,REF7,REF8}, VideoQA is more challenging, which, aside from modeling the semantic correlation between the question and each image, needs to further accurately extract the dynamic interaction between the question and the video.

\begin{figure}[t]
\centering
\includegraphics[width=0.45\textwidth]{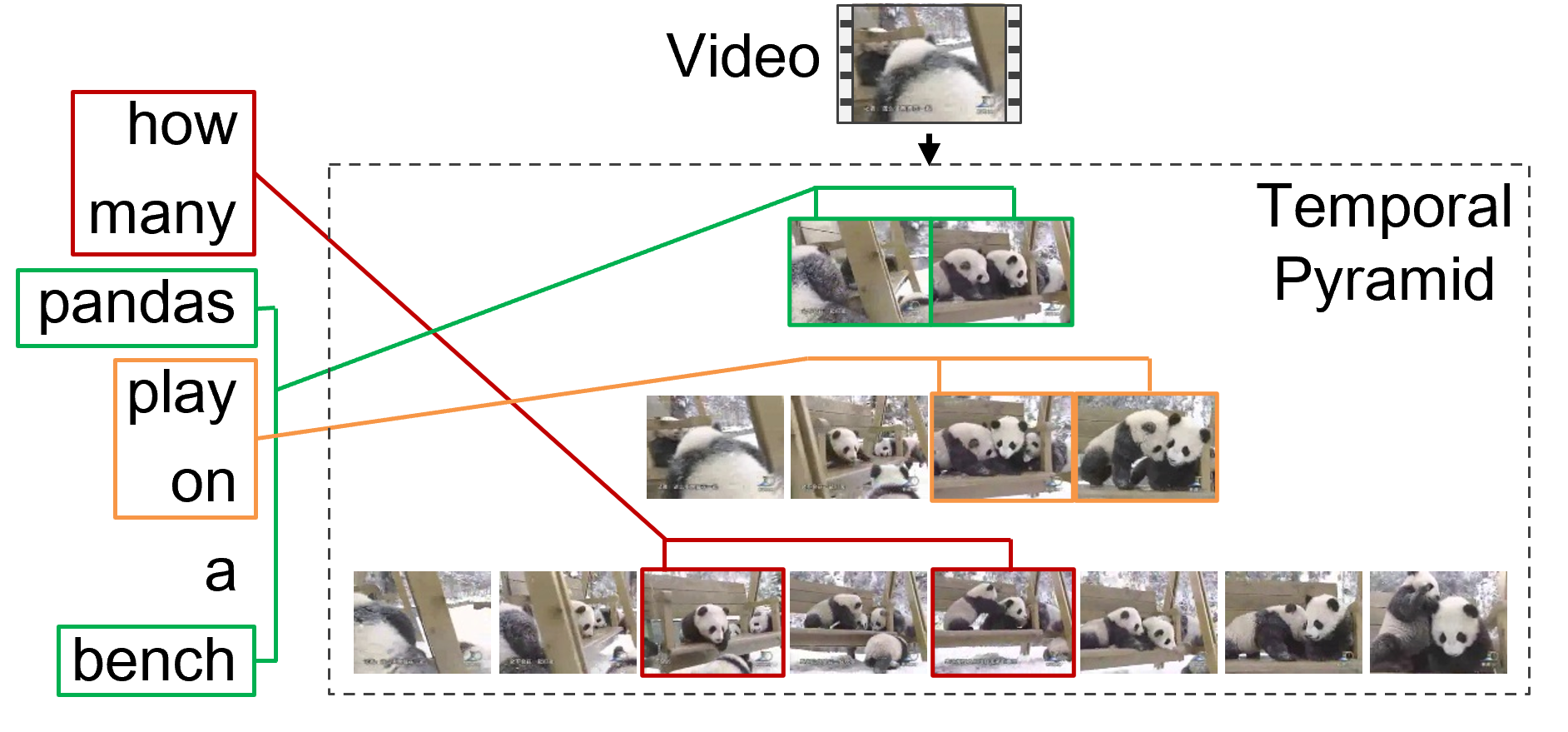}
\caption{The interaction of the question and the visual content usually happens at multiple temporal scales, as illustrated by the connected pairs of different parts of the question and frames at different levels of the temporal pyramid.}
\label{fig1}
\end{figure}

\begin{figure*}[t]
\centering
\includegraphics[width=0.98\textwidth]{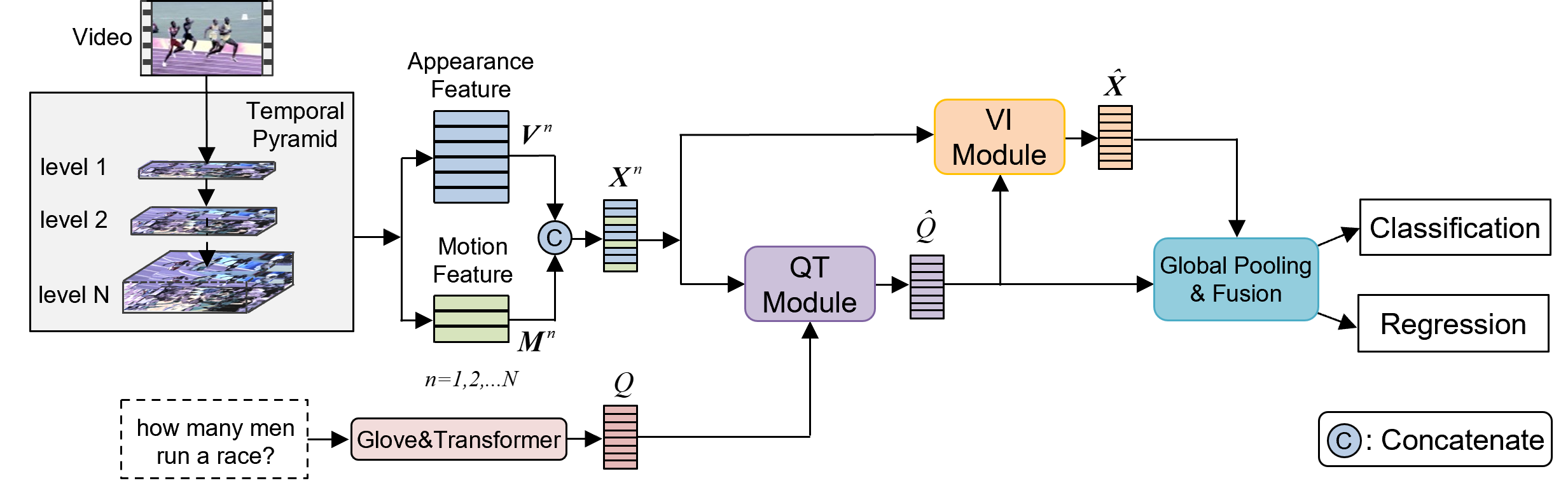}
\caption{An overview of the proposed TPT model. The temporal pyramid is firstly built from the input video sequence, from which the appearance-motion features are extracted at different temporal scales. The QT module adopts a pyramid transformer to extract the multimodal interaction between the question and the appearance-motion feature, to establish a deep understanding of the question. The VI module, guided by the question semantics, builds the local-to-global visual inference from the visual features at different levels.}
\label{fig2}
\end{figure*}

Most existing methods \cite{REF10,REF12,REF13,REF15,REF22,REF24} use recurrent neural networks (RNNs) and their variants to encode the semantic embeddings of the text and spatial features extracted with convolutional neural networks (CNNs) of the video, with spatial-temporal attention mechanisms \cite{REF9,REF16,REF17,REF30} adopted to learn the appearance-motion representations or memory-augmented RNN \cite{REF18,REF25} designed to capture the semantic relation between the text and the video dependent on a larger temporal scale. While these methods achieved competitive results on benchmark datasets, the multi-level interaction between the text and appearance-motion information of the video is largely ignored. Additionally, a deep understanding of the question lacks in studies that used co-attention representation \cite{REF11,REF23,REF24,REF27} and memory-augmented co-representation \cite{REF18,REF22}, which could further influence the learning of respective visual information. The clue seen from the question has a clear hierarchical structure within the visual information, which suggests the need to distill information at different temporal scales from the video. In the example shown in Fig.\ref{fig1}, for understanding the semantics of 'pandas' and 'bench', the informative visual content is found at a coarser level of the temporal pyramid where the global information is sufficient. As the temporal scale increases along the pyramid, visual contents comprising fine-grained and consecutive information are revealed, which are helpful to understand the semantics of 'play on' and 'how many' in the question. This example again suggests that the provision of visual contents at different temporal scales contributes to the interaction between the question and the video, for which the use of temporal pyramid could be a practical approach.

Given the above findings, we propose a novel method named Temporal Pyramid Transformer (TPT) model with multimodal interaction for VideoQA. An overview of the model is shown in Fig.\ref{fig2}. Our TPT model comprises two modules of Question-specific Transformer (QT) and Visual Inference (VI). For the temporal pyramid constructed from a video, based on the feature sequences at different levels, QT extracts the coarse-to-fine semantic co-occurrence between each word and the video, to acquire the semantic understanding of the question in relation to the video. Thereon, VI extracts the local-to-global compositional semantics between the question and the video, and gradually acquire the visual clue needed for answering. Within these two modules, the transformer architecture is extended to adopt an intra-level multimodal attentional interaction scheme, and inter-level residual connections for information passing. In general, integrating the local information at a fine-grained level and the global information at a coarse level, our model leverages the appearance-motion information of different temporal scales to understand the interaction of the text and the video. Experiments on public benchmark datasets demonstrate the improved performances of our method. Our contributions are three-fold:
\begin{itemize}
    \item We propose a novel Temporal Pyramid Transformer (TPT) model with multimodal interaction for VideoQA, which enables interaction of the question and the video at different temporal scales.
    \item We extend the vanilla transformer to adopt a multimodal attention mechanism within each level and residual connections across different levels. 
    \item Through comprehensive experiments on TGIF-QA, MSRVTT-QA, and MSVD-QA datasets, we demonstrate better performances of our method in comparison with the state-of-the-arts.
\end{itemize}

\section{Related Work}

\subsubsection{VideoQA.} The key to VideoQA is to sort out the complex interaction between the question and the video. \cite{REF9} proposed a method based on Gradually Refined Attention to extract the appearance-motion features using the question as guidance. \cite{REF19} proposed a dual-LSTM approach together with spatio-temporal attention to extract video features for VideoQA. Later on, other attention-based methods proposed to use hierarchical attention network \cite{REF20,REF21,REF28}, co-attention representation \cite{REF11,REF23,REF24,REF27}, and memory-augmented co-attention models \cite{REF18,REF22} for the extraction of motion-appearance features and question-related interactions. Recently, \cite{REF29} proposed a multistep progressive attention model to prune out irrelevant temporal segments, and a memory network to progressively update the clue to answer. Additionally, some proposed to leverage object detection across the video to acquire fine-grained appearance-question interactions \cite{REF14,REF26}. \cite{REF31,REF32,REF33} proposed to use hierarchical structures for the extraction of question-video interactions from the frame-level and segment-level. \cite{REF34} proposed a multimodal graph structure using the question graph as an intermediate bridge to extract the internal semantic relation between each word and the video. Although competitive results are achieved by these methods, the hierarchical structure of the answering clue existed in the visual information is not explored.

\subsubsection{Transformer.} As first introduced for neural machine translation, transformer \cite{REF2} has achieved outstanding performance using its self-attention mechanism and forward-passing architecture. Given its efficiency in analyzing temporal information, transformer becomes the dominant approach for various NLP tasks \cite{REF35,REF36}. Recent efforts are seen on introducing transformer into the computer vision domain, for applications in object detection \cite{REF37}, instance segmentation \cite{REF38}, and action recognition \cite{REF39}. In this paper, we first use a transformer architecture to extract semantic embeddings from the question given the visual information, and further extract the semantic interaction between the question and the video for visual inference.

\section{Method}
Given a video $\mathcal{V}$ and the question $\mathcal{Q}$, the aim of VideoQA is to infer the correct answer $a^*$ from an answer space $\mathcal{A}$. For open-ended and multi-choice question types, the answer pace $\mathcal{A}$ comprises pre-defined answers and list of candidate answer options, respectively. Usually, VideoQA can be formulated as follows

\begin{equation}
    a^*=\mathop{\arg\max}\limits_{a\in\mathcal{A}}f_\theta(a\mid \mathcal{Q},\mathcal{V}),
\label{equa1}
\end{equation}

\noindent where $f$ is the mapping function of a model with trainable parameter $\theta$. As shown in Fig.\ref{fig2}, we first build the temporal pyramid from the input video, and extract the appearance-motion features at different levels. Meanwhile, embeddings for the input question is produced. Thereon, we propose a QT module to leverage the question embeddings as queries, to extract the multimodal interaction from the visual features at different levels with a pyramid transformer. As such, we aim to enable a deep understanding of the question embeddings given the relevant visual content. Then, we propose a VI module to finish the visual inference under the guidance of the question semantics. In the end, we combine the features from the two modules to form the final answer output in a classification or regression manner.

\subsection{Temporal Pyramid and Feature Extraction}
Previous works largely ignored the interaction between the text and the video at different temporal scales. In this work, for each input video, we first construct a temporal pyramid that has levels ranging from $1$ to $N$. At different levels, the visual information is stored in different temporal scales. To construct such a temporal pyramid, at $n$-th level, we split the video input into $2^{n-1}$ segments, where each segment comprises $T$ frames consecutively sampled from the video following the original temporal order. That is, different levels of the temporal pyramid has different numbers of video segments sampled from the input video. At level $n$, the group of video segments $\mathbf{S}^n$ can be represented as

\begin{equation}
    \mathbf{S}^n=\lbrace \mathbf{s}^{n,1}, \mathbf{s}^{n,2},...,\mathbf{s}^{n,2^{n-1}} \rbrace,
\label{equa2}
\end{equation}
\noindent with
\begin{equation}
    \mathbf{s}^{n,i}=\lbrace \mathbf{I}^{n,i}_1, \mathbf{I}^{n,i}_2,...,\mathbf{I}^{n,i}_T \rbrace^{2^{n-1}}_{i=1},
\label{equa3}
\end{equation}

\noindent where $\mathbf{I}^{n,i}_1$ is the first frame of $i$-th video segment at level $n$.

A pre-trained ResNet model \cite{REF1} is used to extract the appearance feature from such temporal pyramid. For the group of video segments $\mathbf{S}^n$ at $n$-th level of the temporal pyramid, the group of extracted frame-wise appearance features $\mathbf{V}^n$ can be represented as

\begin{equation}
    \mathbf{V}^n=\lbrace \mathbf{v}^{n,i} \mid \mathbf{v}^{n,i} \in \mathcal{R}^{T\times2048} \rbrace^{2^{n-1}}_{i=1},
\label{equa4}
\end{equation}
\noindent with

\begin{equation}
    \mathbf{v}^{n,i}=\lbrace v^{n,i}_t \mid v^{n,i}_t\in \mathcal{R}^{2048} \rbrace^T_{t=1},
\label{equa5}
\end{equation}

\noindent where each appearance feature vector $v^{n,i}_t$ is extracted from the respective frame $\mathbf{I}^{n,i}_t$. Similarly, with a pre-trained 3D ResNet model \cite{REF3}, we extract the group of segment-wise motion feature $\mathbf{M}^n$ as

\begin{equation}
    \mathbf{M}^n=\lbrace m^{n,i} \mid m^{n,i} \in \mathcal{R}^{2048} \rbrace ^{2^{n-1}}_{i=1},
\label{equa6}
\end{equation}

\noindent where each motion feature vector $m^{n,i}$ is extracted from the respective video segment $\mathbf{s}^{n,i}$.

We use a linear feature transformation layer to map the appearance-motion feature extracted at each level, namely $\mathbf{V}^n$ and $\mathbf{M}^n$, into a $d$-dimensional feature space. Therein, we have $v^{n,i}_t,m^{n,i}\in\mathcal{R}^{d}$. After a feature concatenation following the order of video segments in each level, the appearance-motion feature $\mathbf{X}^n$ at $n$-th level can be represented as

\begin{equation}
    \mathbf{X}^n=\lbrace x^{n}_j \mid x^{n}_j \in \mathcal{R}^{d} \rbrace ^{L^n_X}_{j=1},
\label{equa7}
\end{equation}

\noindent where $x^{n}_j\in\lbrace v^{n,i}_t,m^{n,i} \rbrace$, with total number of appearance-motion features at level $n$ being $L^n_X=2^{n-1}(T+1)$.

For the question and answer candidates, we adopt Glove word embedding method \cite{REF40} to acquire the $300$-dimensional feature embeddings, which is further mapped into a $d$-dimensional space using linear transformation layers. Thereon, a transformer model \cite{REF2} is adopted to extract the contextual semantic information between each word in the question, and the answer, respectively. Finally, the acquired representation for the question and answer candidates can be written as

\begin{equation}
    \mathcal{Q}=\lbrace q_j \mid q_j \in \mathcal{R}^d \rbrace^{L_Q}_{j=1},
\label{equa8}
\end{equation}

\noindent and

\begin{equation}
    \mathcal{A}^k=\lbrace a^k_j \mid a^k_j \in \mathcal{R}^d \rbrace^{L_A^k}_{j=1},
\label{equa9}
\end{equation}

\noindent where $L_Q$ and $L_A^k$ are the number of words in the question and $k$-th answer candidate, respectively.

\subsection{Question-specific Transformer}

\begin{figure}[t]
\centering
\includegraphics[width=0.45\textwidth]{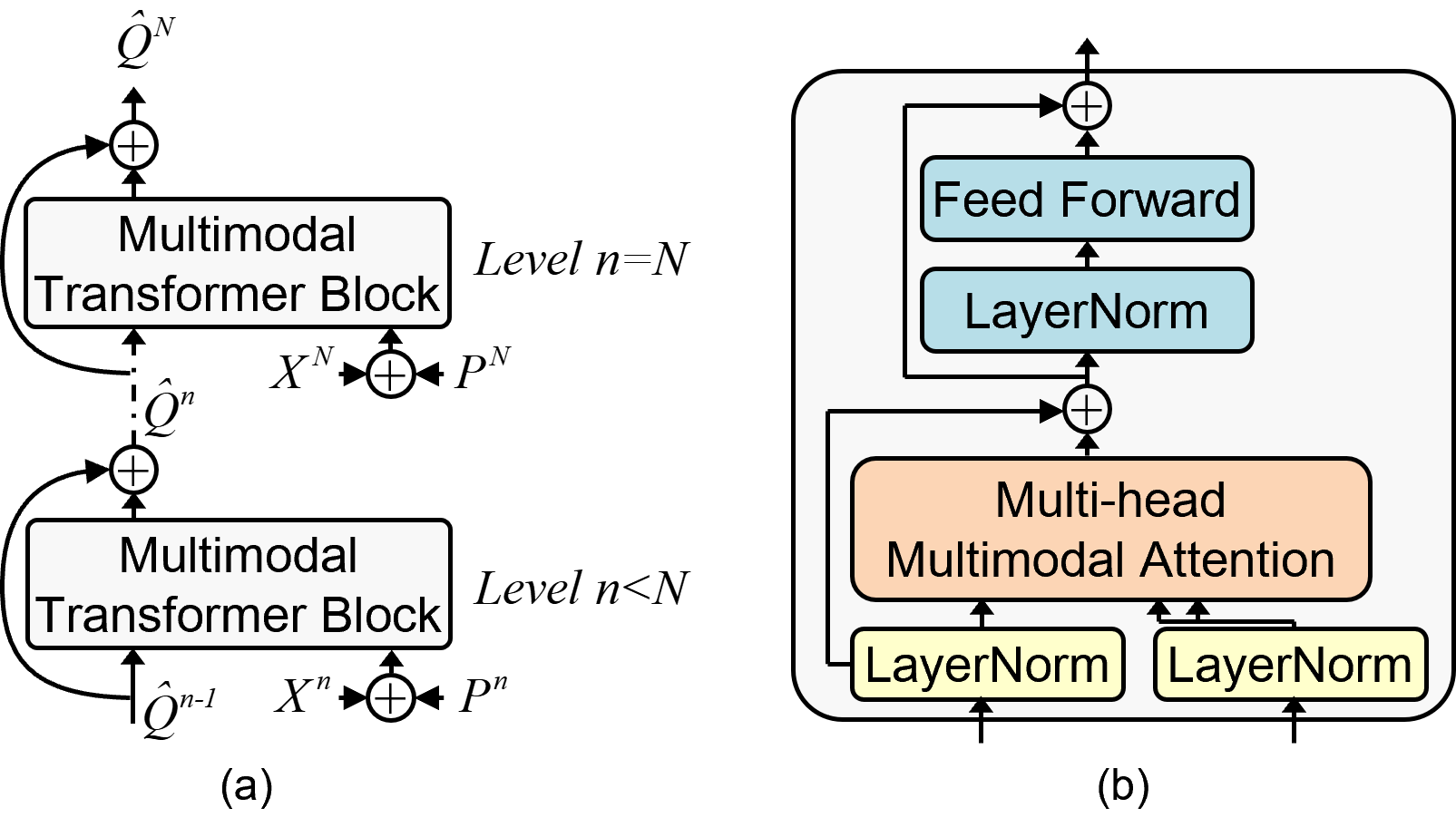}
\caption{(a) The pyramid of multimodal transformer blocks in the QT module. At level $n$, the block comprises a stack of $R$ multimodal transformer layers, each as shown in (b).}
\label{fig3}
\end{figure}

Here, we introduce the QT module, which aims to build a deep understanding of the question in relation to the visual content. As shown in Fig.\ref{fig3}, the QT module comprises a pyramid transformer structure for the multimodal interaction between the question and the appearance-motion features at different levels. Specifically, at each level, a multimodal transformer block is designed comprising $R$ multimodal transformer layers. Such block is also adopted in the VI module, as described in the next subsection. In order to acquire accurate understanding about the question, the output at a lower level is connected with the next level via an external residual connection, as to leverage the fine-grained visual information at higher levels. This also enables the acquisition of video-relevant question semantics. At level $n$, the appearance-motion feature $\mathbf{X}^n$ is added with the learned positional embedding $\mathbf{P}^n\in\mathcal{R}^{L_X^n \times d}$, in order to maintain the positional information of the video feature sequence. Thereon, for $r=1,2,...,R$ layers of the transformer block at this level, the multimodal output given semantics representation $\hat{\mathcal{Q}}^{n-1}$ and $\mathbf{X}^n$ is computed as

\begin{equation}
    \hat{\mathcal{Q}}^{n-1,r}=\mathbf{Z}^{n,r}+\sigma(\mathrm{LN}(\mathbf{Z}^{n,r})),
\label{equa10}
\end{equation}

\noindent with

\begin{equation}
    \mathbf{Z}^{n,r}=\mathrm{LN}(\hat{\mathcal{Q}}^{n-1,r-1})+\mathrm{MMA}(\hat{\mathcal{Q}}^{n-1,r-1},\mathbf{X}^n),
\label{equa11}
\end{equation}

\noindent where $\hat{\mathcal{Q}}^{n-1,0}=\hat{\mathcal{Q}}^{n-1}$, $\hat{\mathcal{Q}}^{0,0}=\mathcal{Q}$. $\mathbf{Z}^{n,r}$ is the intermediate feature at $r$-th layer of the block at level $n$. $\mathrm{LN}(\cdot)$ is the layer normalization operation \cite{REF41}, $\sigma(\cdot)$ is the feed forward layer that comprises two linear projections separated by a GELU non-linear activation. The dimension $d$ of each feature matrix stays unchanged across all layers at the same level. $\mathrm{MMA}(\cdot)$ is the Multi-head Multimodal Attention layer, with the output at an attention head $h$ being

\begin{equation}
    \mathrm{MMA}^{n,h}(\hat{\mathcal{Q}}^{n-1,r-1},\mathbf{X}^n)=\mathrm{softmax}(\frac{\mathbf{F}_q^h\mathbf{F}_k^{h\mathrm{\top}}}{\sqrt{d}})\mathbf{F}_v^h,
\label{equa12}
\end{equation}

\noindent where $\mathbf{F}_q^h=\mathrm{LN}(\hat{\mathcal{Q}}^{n-1,r-1})\mathbf{W}_q^h$ is the Query, $\mathbf{F}_k^h=\mathrm{LN}(\mathbf{X}^n)\mathbf{W}_k^h$ is the Key, and $\mathbf{F}_v^h=\mathrm{LN}(\mathbf{X}^n)\mathbf{W}_v^h$ is the Value. The final output of $\mathrm{MMA}$ at each level $n$ is the concatenation of the feature computed per head, written as

\begin{equation}
    \mathrm{MMA}^n=\mathrm{concat}(\mathrm{MMA}^{n,1},...,\mathrm{MMA}^{n,H})\mathbf{W}_o,
\label{equa13}
\end{equation}

\noindent where $\mathbf{W}_q^h$, $\mathbf{W}_k^h$, $\mathbf{W}_v^h\in\mathcal{R}^{d\times d/H}$, and $\mathbf{W}_o\in\mathcal{R}^{d\times d}$ are the learnable weight matrices, and $H$ is the total number of attention heads. In short, within the $\mathrm{MMA}$, the question semantics and appearance-motion features are connected, while the semantic co-occurrence of them is extracted using the attention mechanism. Such attentional output is further added with the question semantics via an internal residual connection to form the visual-relevant textual semantic representation across different layers of the transformer architecture at each level (Equation\ref{equa10}). Given the external residual connection, the question semantic output $\hat{\mathcal{Q}}^n$ at $n$-th level is computed as

\begin{equation}
    \hat{\mathcal{Q}}^n=\hat{\mathcal{Q}}^{n-1,R}+\hat{\mathcal{Q}}^{n-1}.
\label{equa14}
\end{equation}

Finally, at the $N$-th level of the QT pyramid, the output for such pyramid is $\hat{\mathcal{Q}}=\hat{\mathcal{Q}}^N$.

\begin{figure}[b]
\centering
\includegraphics[width=0.25\textwidth]{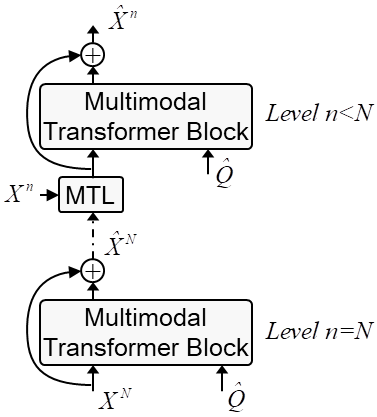}
\caption{The pyramid of multimodal transformer blocks in the VI module. At level $n<N$, the visual input to the block is first processed by a single multimodal transformer layer.}
\label{fig4}
\end{figure}

\subsection{Visual Inference}

An essential step of VideoQA is to infer the visual clue within the appearance-motion feature given the semantics of the question. Within the VI module, as shown in Fig.\ref{fig4}, a similar pyramid of multimodal transformer blocks is adopted. In comparison to the QT module, given the question semantics representation $\hat{\mathcal{Q}}$, we search the clue to answering across different temporal scales of the appearance-motion feature $\mathbf{X}^n$. For the pyramid, the representation extracted at a higher level has richer local information, while at a lower level it possesses more complete global information. Here, we conduct a local-to-global search of the visual clues, by connecting the blocks with residual connections from level $N$ to level $1$.

In detail, at the level $n=N$ of the pyramid and similar to Equation\ref{equa10} and \ref{equa11}, the multimodal interaction output given the appearance-motion feature $\mathbf{X}^N$ and question semantics input $\hat{\mathcal{Q}}$ is computed as

\begin{equation}
    \mathbf{X}^{N,r}=\mathbf{Z}^{N,r-1}+\sigma(\mathrm{LN}(\mathbf{Z}^{N,r-1})),
\label{equa15}
\end{equation}

\noindent with

\begin{equation}
    \mathbf{Z}^{N,r-1}=\mathbf{X}^{N,r-1}+\mathrm{MMA}(\mathbf{X}^{N,r-1},\hat{\mathcal{Q}}),
\label{equa16}
\end{equation}

\noindent where each transformer block comprises $r=1,2,...,R$ layers, and $\mathbf{X}^{N,0}=\mathbf{X}^{N}$. Given the external residual connection, the visual clue output $\hat{\mathbf{X}}^{N}$ at $N$-th level is computed as

\begin{equation}
    \hat{\mathbf{X}}^{N}=\mathbf{X}^{N,R}+\mathbf{X}^{N}.
\label{equa17}
\end{equation}

From the level $n<N$, we use a single Multimodal Transformer Layer (MTL) to fuse the visual output $\hat{\mathbf{X}}^{n+1}$ of the last level with the current visual input $\mathbf{X}^n$

\begin{equation}
    \tilde{\mathbf{X}}^n=\mathcal{X}^n+\sigma(\mathrm{LN}(\mathcal{X}^n)).
\label{equa18}
\end{equation}

\noindent with

\begin{equation}
    \mathcal{X}^n=\hat{\mathbf{X}}^{n+1}+\mathrm{MMA}(\mathbf{X}^n,\hat{\mathbf{X}}^{n+1}).
\label{equa19}
\end{equation}

Then, similar to Equation\ref{equa15}, \ref{equa16}, and \ref{equa17}, the output of MTL, namely the $\tilde{\mathbf{X}}^n$ and question semantics representation $\hat{\mathcal{Q}}$, is processed by a multimodal transformer block to acquire the visual clue output $\hat{\mathbf{X}}^n$ at this level.

Finally, as shown in Fig.\ref{fig2}, an average pooling is applied for the temporal dimension separately on the question semantics representation $\hat{\mathcal{Q}}$ produced by QT module and the visual clue representation $\hat{\mathbf{X}}$ produced by VI module. Then the two feature matrices are concatenated to make the output of TPT model $\bar{\mathbf{O}}\in\mathcal{R}^{2d}$.

\subsection{Answer Decoding and Loss Computation}
Following previous studies \cite{REF11,REF13,REF14,REF18,REF32,REF34}, different decoding methods are used according to the types of question. Specifically, we approach an open-ended question as a multi-class classification task, where the answering decoder aims to predict the correct class from the answer space $\mathcal{A}$. Given the final output $\bar{\mathbf{O}}$ of the TPT model, the probability vector $\mathbf{P}\in\mathcal{R}^{|\mathcal{A}|}$ towards each class is computed as 

\begin{equation}
    y=\delta(\mathbf{W}^o\bar{\mathbf{O}}+b^o),
\label{equa20}
\end{equation}

\begin{equation}
    \mathbf{P}=\mathrm{softmax}(\mathbf{W}^yy+b^y),
\label{equa21}
\end{equation}

\noindent where $\mathbf{W}^o$, $\mathbf{W}^y$, $b^o$, and $b^y$ are the learnable weight matrices and biases of each layer, respectively; $\delta(\cdot)$ is the activation function. The cross-entropy loss is used here.

For the repetition count task, linear regression is adopted to replace the classification function shown in Equation\ref{equa21}, the output of which is processed by a rounding function to acquire the integer output. The loss function used here is the Mean Squared Error (MSE).

For multi-choice question types, we use an answer candidate $\mathcal{A}^k$ as input to the TPT model, similar with the question $\mathcal{Q}$. Therein, the learnable parameters are shared for the processing of the answer and the question. Given the output $\bar{\mathbf{O}}$ conditioned by the question input and $\bar{\mathbf{O}}^k_a$ conditioned by the $k$-th answer candidate, the predicted probability towards the $k$-th answer candidate is computed as

\begin{equation}
    y^k=\delta(\mathbf{W^*}[\bar{\mathbf{O}};\bar{\mathbf{O}}^k_a]+b^*).
\label{equa22}
\end{equation}

\begin{equation}
    p^k=\mathbf{W}^*y^k+b^*.
\label{equa23}
\end{equation}

The answer candidate that produces the highest probability $p$ is selected as the final answer for the question. Hinge loss \cite{REF42}, namely $\mathrm{max}(0,1+p^k-p^{GT})$, is adopted to compute the loss between the correct answer and the other candidates. $p^{GT}$ is the predicted probability towards the groundtruth answer candidate.

\section{Experiment}

\begin{table*}[t]
\centering
\caption{Comparison with state-of-the-art methods on TGIF-QA dataset. For the Count task, the lower is better. The best performance is marked in bold per task.}
\label{tab:table1}
\resizebox{0.75\textwidth}{!}{%
\begin{tabular}{l|cccc|c}
\hline
Method                     & Action        & Trans.         & FrameQA       & Count $\downarrow$ & Reference \\ \hline
ST-TP \cite{REF19}         & 62.9          & 69.4          & 49.5          & 4.32               & CVPR 17   \\
Co-Mem \cite{REF16}        & 68.2          & 74.3          & 51.5          & 4.10               & CVPR 18   \\
PSAC \cite{REF11}          & 70.4          & 76.9          & 55.7          & 4.27               & AAAI 19   \\
HME \cite{REF18}           & 73.9          & 77.8          & 53.8          & 4.02               & CVPR 19   \\
L-GCN \cite{REF14}         & 74.3          & 81.1          & 56.3          & 3.95               & AAAI 20   \\
QueST \cite{REF13}         & 75.9          & 81.0          & \textbf{59.7} & 4.19               & AAAI 20   \\
HCRN \cite{REF32}          & 75.0          & 81.4          & 55.9          & 3.82               & CVPR 20   \\
Bridge2Answer \cite{REF34} & 75.9          & \textbf{82.6} & 57.5          & 3.71               & CVPR 21   \\ \hline
\textbf{TPT (ours)}     & \textbf{76.6} & 81.6          & 57.8          & \textbf{3.63}      & -         \\ \hline
\end{tabular}
}
\end{table*}

\begin{table*}[t]
\centering
\caption{Comparison with state-of-the-art methods on MSVD-QA and MSRVTT-QA datasets. The best performance is marked in bold per dataset.}
\label{tab:table2}
\resizebox{0.9\textwidth}{!}{%
\begin{tabular}{l|c|c|c}
\hline
Method                     & MSVD-QA \cite{REF9} & MSRVTT-QA \cite{REF43} & Reference \\ \hline
AMU \cite{REF9}            & 32.0                & 32.5                   & MM 17     \\
HRA \cite{REF21}           & 34.4                & 35.0                   & ICIP 18   \\
Co-Mem \cite{REF16}        & 31,7                & 31.9                   & CVPR 18   \\
HME \cite{REF18}           & 33.7                & 33.0                   & CVPR 19   \\
QueST \cite{REF13}         & 36.1                & 34.6                   & AAAI 20   \\
HCRN \cite{REF32}          & 36.1                & 35.6                   & CVPR 20   \\
Bridge2Answer \cite{REF34} & 37.2                & 36.9                   & CVPR 21   \\ \hline
\textbf{TPT (ours)}     & \textbf{38.5}       & \textbf{37.7}          & -         \\ \hline
\end{tabular}
}
\end{table*}

\subsection{Datasets}
Three benchmark datasets for VideoQA are adopted for the evaluation, which comprises videos of various lengths and different question types.

\subsubsection{TGIF-QA} \cite{REF19} is a large-scale dataset for videoQA, which comprises 165K question-answer pairs and 72K animated GIFs. Four types of task are included in this dataset, including \textit{Action}, \textit{Transition} (Trans.), \textit{FrameQA}, and \textit{Count}. Action is a multi-choice task aimed to identify the repetitive actions. Trans. is another multi-choice task for identifying the transition actions before or after a target action. FrameQA is an open-ended task where the answer could be inferred from a single frame of the video (GIF file). Count is a task to count the number of a repetitive action.

\subsubsection{MSVD-QA} \cite{REF9} comprises 1,970 short clips and 50,505 question-answer pairs, which are divided into five question categories of \textit{what}, \textit{who}, \textit{how}, \textit{when}, and \textit{where}. All of these questions are open-ended.

\subsubsection{MSRVTT-QA} \cite{REF43} comprises 10K videos and 243K question-answer pairs. The question types are similar to what included in the MSVD-QA dataset. However, the scenario contained in the video is more complex than the above two datasets, with a longer duration of 10-30 seconds.

\subsection{Implementation Details}

\subsubsection{Metrics.}
For multi-choice and open-ended tasks, we use the accuracy to evaluate the performance of a videoQA method. For the Count task in TGIF-QA dataset, we adopt the Mean Squared Error (MSE) between the predicted answer and the answer to evaluate the performance.

\subsubsection{Training Details.}
For all the experiments, we use the official split of training, validation, and testing sets of each dataset. For each input video, as described in an earlier section, a $N$ level temporal pyramid structure is built, with $N$ set to 3 by default. The number of frames $T$ in each video segment at a level is set to 16 by default. For the multimodal transformer block used in QT and VI modules, the number of layers $R$ is set to 3, the feature dimension $d$ is set to 512, and the number of attentional heads $H$ is set to 8. The number of mini batch size is set to 64. The Adam \cite{REF44} optimizer is adopted, with the initial learning rate set to 1e-4, which reduces by half when the loss stops decreasing after every 5 epochs. We set the number of epochs to 50. We implement the method with PyTorch deep learning library on a PC with two GTX 1080 Ti GPUs and memory of 16 GB.

\subsection{Comparison with the State-of-the-arts}
\begin{figure}[t]
\centering
\includegraphics[width=0.46\textwidth]{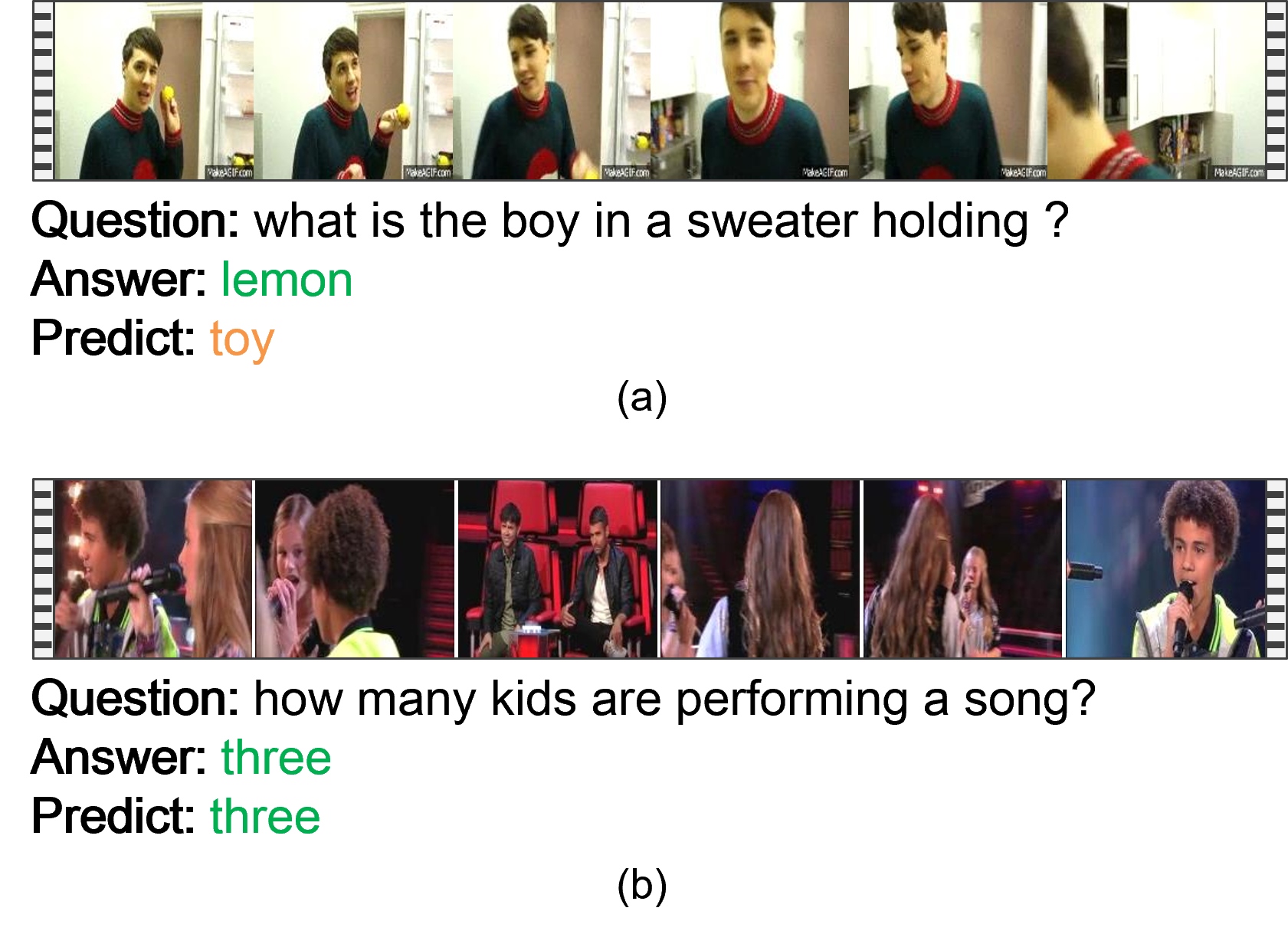}
\caption{Visualizations of the model's performance on video examples taken from (a) TGIF-QA and (b) MSRVTT-QA datasets, respectively.}
\label{fig5}
\end{figure}

On the TGIF-QA dataset, our method is compared with the methods of ST-TP \cite{REF19}, Co-Mem \cite{REF16}, PSAC \cite{REF11}, HME \cite{REF18}, L-GCN \cite{REF14}, QueST \cite{REF13}, HCRN \cite{REF32}, and Bridge2Answer \cite{REF34}. As shown in Table\ref{tab:table1}, our TPT model achieves better results than others on the tasks of Action and Count. The Trans. task particularly relies on the capacity of searching information bidirectionally across the temporal features, for which the modeling of temporal features as graphs seen in \cite{REF34} shows better performance, with our method ranked the 2nd. Additionally, the use of vanilla ResNet for spatial feature extraction in our method poses a certain limitation on capturing detailed spatial information in a single frame, thus a better performance is achieved by \cite{REF13} that leverages a spatial-temporal attention mechanism for fine-grained spatial understanding. A qualitative example is shown in Fig.\ref{fig5} (a), where our method misunderstands the 'lemon' taking up a small spatial area of the frame as the 'toy'.

Further comparisons with AMU \cite{REF9} and HRA \cite{REF21} on the MSVD-QA and MSRVTT-QA datasets are conducted, which are more challenging in terms of scenario complexity and video length. Results are listed in Table\ref{tab:table2}. As shown, our TPT model achieves the best performances of 38.5\% and 37.7\% on both datasets, respectively. We use another video sample (Fig.\ref{fig5} (b)) taken from the MSRVTT-QA dataset to qualitative demonstrate that, given a video with more complex scenario, our model is able to discover informative visual clues appeared at different temporal scales for VideoQA.

\subsection{Ablation Study}
Here, we run several ablation experiments on the TGIF-QA dataset to better understand the impact of different components proposed in our method. We adopt the default TPT model used in the comparison experiment of the last subsection as the basic variant. First, we study the use of the temporal pyramid constructed from the video as the input to the model, in terms of its importance and the impact of different numbers of levels used for different task types. Then, we evaluate the importance of the QT module proposed for better question semantics extraction.

\begin{table}[t]
\centering
\caption{The impact of the temporal pyramid as input to the model on videoQA in the TGIF-QA dataset. For variants without using the temporal pyramid, the input to the model is the scale-fixed output at the specified single level $L$ of the pyramid. TP=Temporal Pyramid.}
\label{tab:table3}
\resizebox{0.46\textwidth}{!}{%
\begin{tabular}{lcccc}
\hline
Model              & Action        & Trans.        & FrameQA       & Count $\downarrow$ \\ \hline
TPT w/o TP, $L$=1 & 74.7          & 76.5          & 57.1          & 4.23               \\
TPT w/o TP, $L$=2 & 75.5          & 78.2          & 57.4          & 3.92               \\
TPT w/o TP, $L$=3 & 76.2          & 79.2          & 57.3          & 3.74               \\
TPT (default)            & \textbf{76.6} & \textbf{81.6} & \textbf{57.8} & \textbf{3.63}      \\ \hline
\end{tabular}%
}
\end{table}

\subsubsection{Impact of the Temporal Pyramid.}
The temporal pyramid is used across the model to establish the multi-level interaction between the appearance-motion feature of the video and the question. Here, we first analyze the impact of using such feature construction strategy. Specifically, while the complete temporal pyramid is removed from the model, the feature output at single levels (i.e., $L=1$, or $L=2$, or $L=3$)
is used as the input for the variants. Results are shown in Table\ref{tab:table3}. Without the temporal pyramid, the input of a single temporal scale leads to decrease in performance across all the tasks of the TGIF-QA dataset. Such reduction is more obvious for the tasks of Action, Trans., and Count, where a better visual clue could be searched using the information at different temporal scales. This suggests the importance of using a temporal pyramid structure for the interaction between the appearance-motion information and the question in a global-to-local manner for such diverse tasks in videoQA.

\begin{table}[b]
\centering
\caption{The impact of different numbers of level $N$ of the temporal pyramid as the input on videoQA in the TGIF-QA dataset.}
\label{tab:table4}
\resizebox{0.46\textwidth}{!}{%
\begin{tabular}{lcccc}
\hline
Parameter $N$ & Action        & Trans.        & FrameQA       & Count $\downarrow$ \\ \hline
$N=1$         & 75.7          & 78.3          & 56.8          & 4.22               \\
$N=2$         & 76.2          & 79.6          & 57.7          & 3.98               \\
$N=3$         & 76.6 & \textbf{81.6} & \textbf{57.8} & 3.63               \\
$N=4$         & 76.6          & 80.2          & 57.7          & \textbf{3.62}      \\ \hline
\end{tabular}%
}
\end{table}

\subsubsection{Impact of the Number of Pyramid Levels.}
The number of levels $N$ of the temporal pyramid determines the \textit{depth} of the interaction between the video content and the question. At a higher level, the number of video frames is bigger so that the interaction becomes fine-grained. Therein, however, the information redundancy and noise inference become more obvious. On the other hand, at a lower level, the appearance-motion feature preserves better global information. However, the local information is lost therein, which could be harmful for the task like Count in the TGIF-QA dataset. Here, we change the number of levels $N$ used for the temporal pyramid as input to the model. Results are shown in Table\ref{tab:table4}. As shown, the performance of the model on different tasks reacts differently towards the increase of the number of pyramid levels $N$. The performance on FrameQA fluctuates very little with the increase of pyramid levels, which may suggest that the global information at lower levels is sufficient for the task. The performance on Count is improved, as finer appearance-motion feature is acquired with more frames available at higher levels. Such is also the case for the Action task. The performance on the Trans. task, on the other hand, seems to be more dependent on the balance between insufficient visual information and redundancy. In general, the visual information provided by the temporal pyramid at different levels offers visual clues with different granularity. Thus, for different tasks, a match between the number of levels and the question type should be considered.

\subsubsection{Impact of the QT module.}

\begin{table}[t]
\centering
\caption{The impact of the QT module for question understanding on VideoQA in the TGIF-QA dataset.}
\label{tab:table5}
\resizebox{0.46\textwidth}{!}{
\begin{tabular}{lcccc}
\hline
Model           & Action        & Trans.         & FrameQA       & Count. $\downarrow$ \\ \hline
TPT w/o QT & 76.2          & 79.9          & 57.5          & 3.67               \\
TPT (default)              & \textbf{76.6} & \textbf{81.6} & \textbf{57.8} & \textbf{3.63}      \\ \hline
\end{tabular}%
}
\end{table}

The QT module is proposed to enable an accurate understanding of the textual question in relation to the visual content, which is essential for videoQA. In this experiment, the QT module is removed from the model, while the question semantic output $\hat{\mathcal{Q}}$ is replaced with the embedded question representation $\mathcal{Q}$ directly. As shown in Table\ref{tab:table5}, the performance of the model drops for all the tasks in the TGIF-QA dataset. Such result suggests the importance of building semantic understanding of the question in relation to the visual input, while the feature embedding produced solely from the question itself is less informative.

\section{Conclusion}
In this paper, we proposed a novel Temporal Pyramid Transformer (TPT) model with multimodal interaction for videoQA, which comprises QT and VI modules for better question understanding and visual inference, respectively. In detail, under the guidance of the original question embeddings, the QT module acquires the visual-relevant question semantics from the temporal pyramid of the appearance-motion features. Thereon, the VI module further searches the visual clue from the feature pyramid. For both modules, we proposed a multimodal transformer architecture to help extract the multi-level interaction information, and used residual connections to communicate the information across different levels of the respective pyramid. Comparison experiments on three benchmark datasets demonstrated the advantage of our method against the state-of-the-arts. While the ablation experiments generally justified the overall design of the model, we particularly found that the number of levels used for the pyramid shall be adjusted according to the task type, as to reach a balance between insufficient visual information and redundancy. 

Future works may try to improve the capacity of the appearance feature extraction of the model for better capturing small targets in the image. Additionally, one may modify the temporal processing strategy, e.g. to enable a bidirectional search, for informative events that could exist at different temporal positions of the same scale.

% Use \bibliography{yourbibfile} instead or the References section will not appear in your paper
\bibliography{references}

\section{Acknowledgments}
This work is funded by the National Natural Science Foundation of China (Grant No. 61806185). Chongyang Wang is supported by the UCL Overseas Research Scholarship (ORS) and Graduate Research Scholarship (GRS).

\end{document}